\documentclass[10pt,twocolumn,letterpaper]{article}

\usepackage[pagenumbers]{cvpr} 

\usepackage{amsmath}
\usepackage{graphicx}
\usepackage{booktabs}
\usepackage{xcolor}
\usepackage{colortbl}
\usepackage{algorithm}
\usepackage{algpseudocode}
\usepackage{rotating}
\usepackage{caption} 
\usepackage{multirow}
\DeclareUnicodeCharacter{200B}{}
\usepackage{pgfplots}  
\pgfplotsset{compat=1.17} 
\usepackage{tabularx}
\usepackage{array}

\definecolor{cvprblue}{rgb}{0.21,0.49,0.74}
\usepackage[pagebackref,breaklinks,colorlinks,allcolors=cvprblue]{hyperref}

\def\paperID{12809} 
\def\confName{CVPR}
\def\confYear{2026}

\title{FlowSteer: Guiding Few-Step Image Synthesis with Authentic Trajectories}

\author{
    Lei Ke$^{1}$ \quad Hubery Yin$^{2}$ \quad Gongye Liu$^{3}$ \quad Zhengyao Lv$^{4}$ \quad Jingcai Guo$^{5}$ \\
    Chen Li$^{2}$ \quad Wenhan Luo$^{3}$ \quad Yujiu Yang$^{1}$\thanks{Corresponding author.} \quad Jing Lyu$^{2}$ \\[1.5ex]
    $^{1}$Tsinghua University \quad $^{2}$WeChat Vision, Tencent Inc. \\
    $^{3}$The Hong Kong University of Science and Technology \\
    $^{4}$The University of Hong Kong \quad $^{5}$The Hong Kong Polytechnic University \\
    {\tt\small kl23@mails.tsinghua.edu.cn, yang.yujiu@sz.tsinghua.edu.cn}
}

\begin{document}
\maketitle
\begin{abstract}

\noindent 
With the success of flow matching in visual generation, sampling efficiency remains a critical bottleneck for its practical application. 
Among flow models' accelerating methods, ReFlow has been somehow overlooked although it has theoretical consistency with flow matching. This is primarily due to its suboptimal performance in practical scenarios compared to consistency distillation and score distillation. 
In this work, we investigate this issue within the ReFlow framework and propose FlowSteer, a method unlocks the potential of ReFlow-based distillation by guiding the student along teacher's authentic generation trajectories. 
We first identify that Piecewised ReFlow's performance is hampered by a critical distribution mismatch during the training and propose Online Trajectory Alignment(OTA) to resolve it. 
Then, we introduce a adversarial distillation objective applied directly on the ODE trajectory, improving the student's adherence to the teacher's generation trajectory. 
Furthermore, we find and fix a previously undiscovered flaw in the widely-used \texttt{FlowMatchEulerDiscreteScheduler} that largely degrades few-step inference quality. Our experiment result on SD3 demonstrates our method's efficacy.

\end{abstract}    
\section{Introduction}
\label{sec:introduction}

Flow matching~\cite{flowmatch, flowstraight} has rapidly established itself as a leading paradigm in visual generation, surpassing traditional diffusion models~\cite{ddpm, edm} in synthesizing high-fidelity content. Its core strength lies in learning straight, efficient generation trajectories, which theoretically enables high-quality synthesis in few steps. However, due to the complex data mapping, the learned trajectories are rarely perfectly linear, flow models still face inference efficiency issue.

To address this, many acceleration techniques have been developed. Among them, ReFlow~\cite{instaflow} aims to straighten models' ODE trajectory to achieve few-step inference. Building on this principle, Piecewise Rectified Flow (PeRFlow)~\cite{perflow} further enhances few-step generation by employing a divide-and-conquer strategy. It partitions the full generation trajectory into multiple shorter stages and straightens each segment individually.
Although it has theoretical consistency with flow matching, ReFlow-based methods~\cite{instaflow, perflow, rectifieddiffusion} have been somewhat overlooked by the community because of their suboptimal performance in practical applications.
In this work, we diagnose a critical distribution issue within PeRFlow through both theoretical analysis and empirical evidence. We then introduce Online Trajectory Alignment(OTA) to resolve it. Furthermore, we apply adversarial distillation directly on the teacher-and-student trajectories to further enhance performance.

\begin{figure}[t!]
    \vspace{-40pt}
    \centering
    \begin{subfigure}[b]{0.46\columnwidth}
        \centering
        \includegraphics[trim=50pt 220pt 35pt 80pt, clip, width=1.1\linewidth]{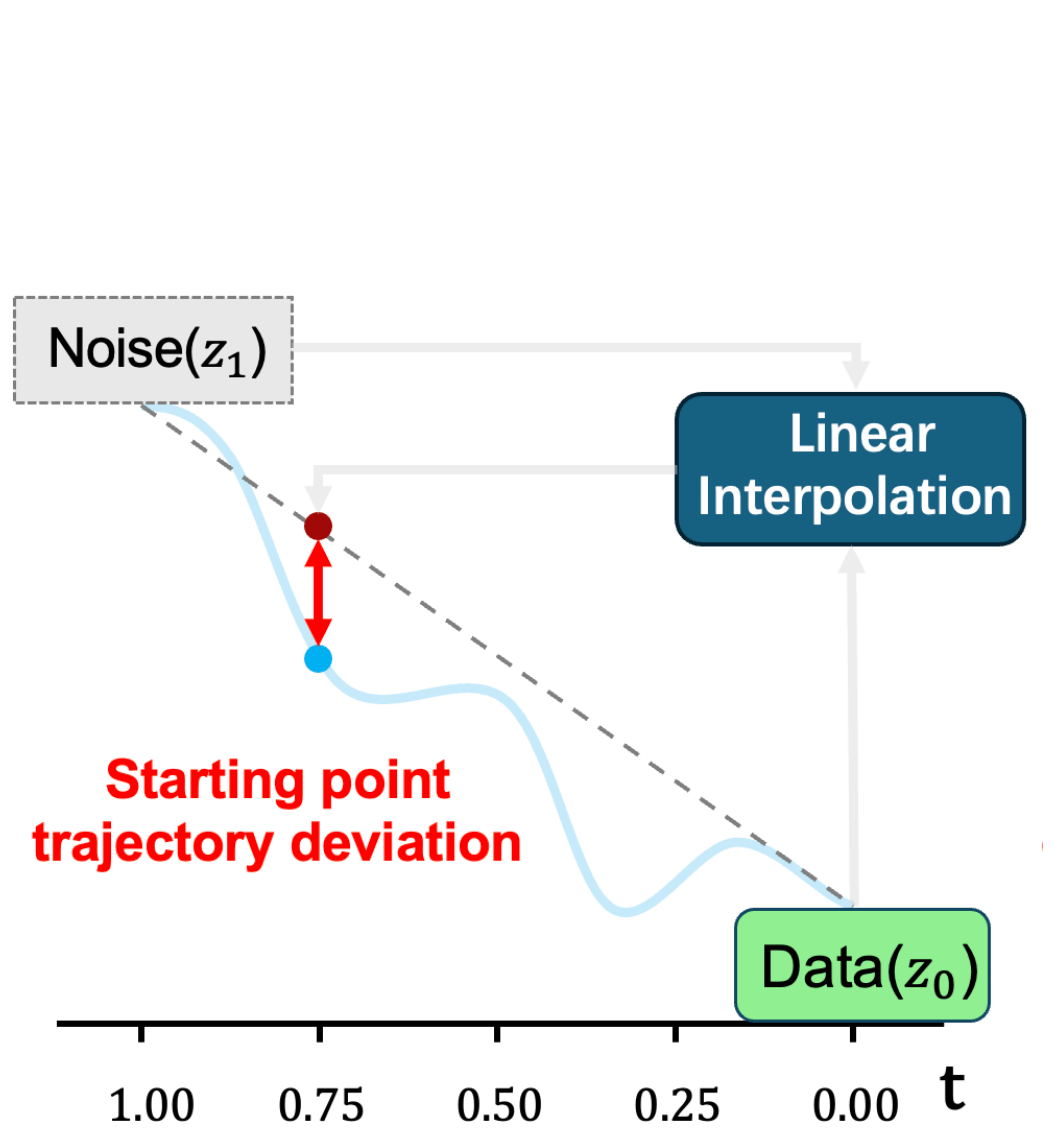}
        \caption{Teacher Trajectory Mismatch.}
        \label{fig:mismatch_a}
    \end{subfigure}
    \hfill 
    \begin{subfigure}[b]{0.52\columnwidth}
        \centering
        \includegraphics[trim=25pt 220pt 25pt 70pt, clip, width=1.05\linewidth]{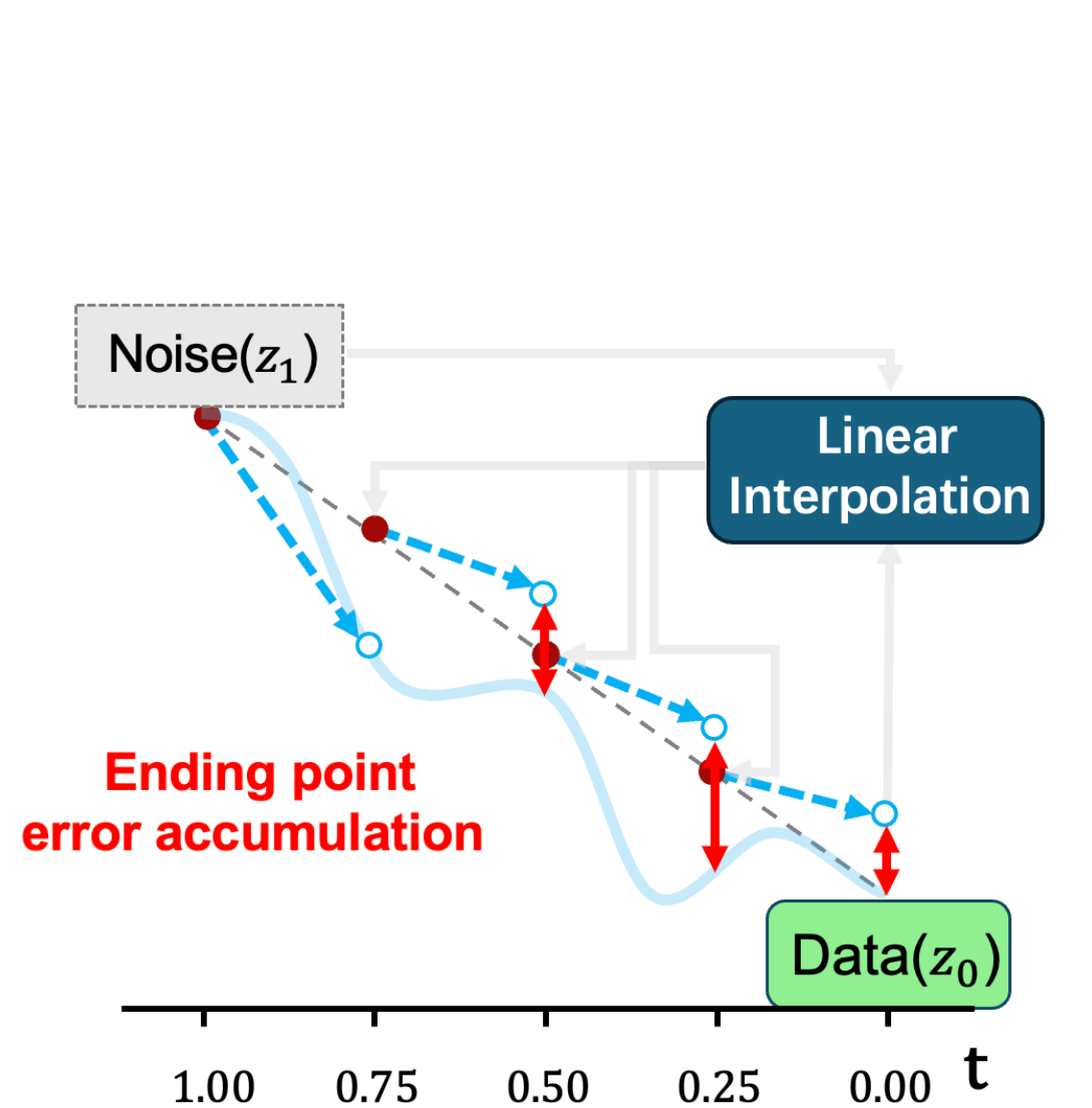}

        \caption{Inter-Stage Distribution Mismatch.}
        \label{fig:mismatch_b}
    \end{subfigure}
    
    \caption{
        Two key mismatches caused by off-trajectory training. 
        \textbf{(a) Teacher Trajectory Mismatch:} The teacher's trajectory deviates from its own generation process. 
        \textbf{(b) Inter-Stage Distribution Mismatch:} Linear interpolation introduces distribution mismatch between consecutive stages, leading to error accumulation.
    }
    \label{mismatch}
\end{figure}

\begin{figure}[t!] 
    \centering 
    \vspace{-20pt}
    \includegraphics[width=\linewidth]{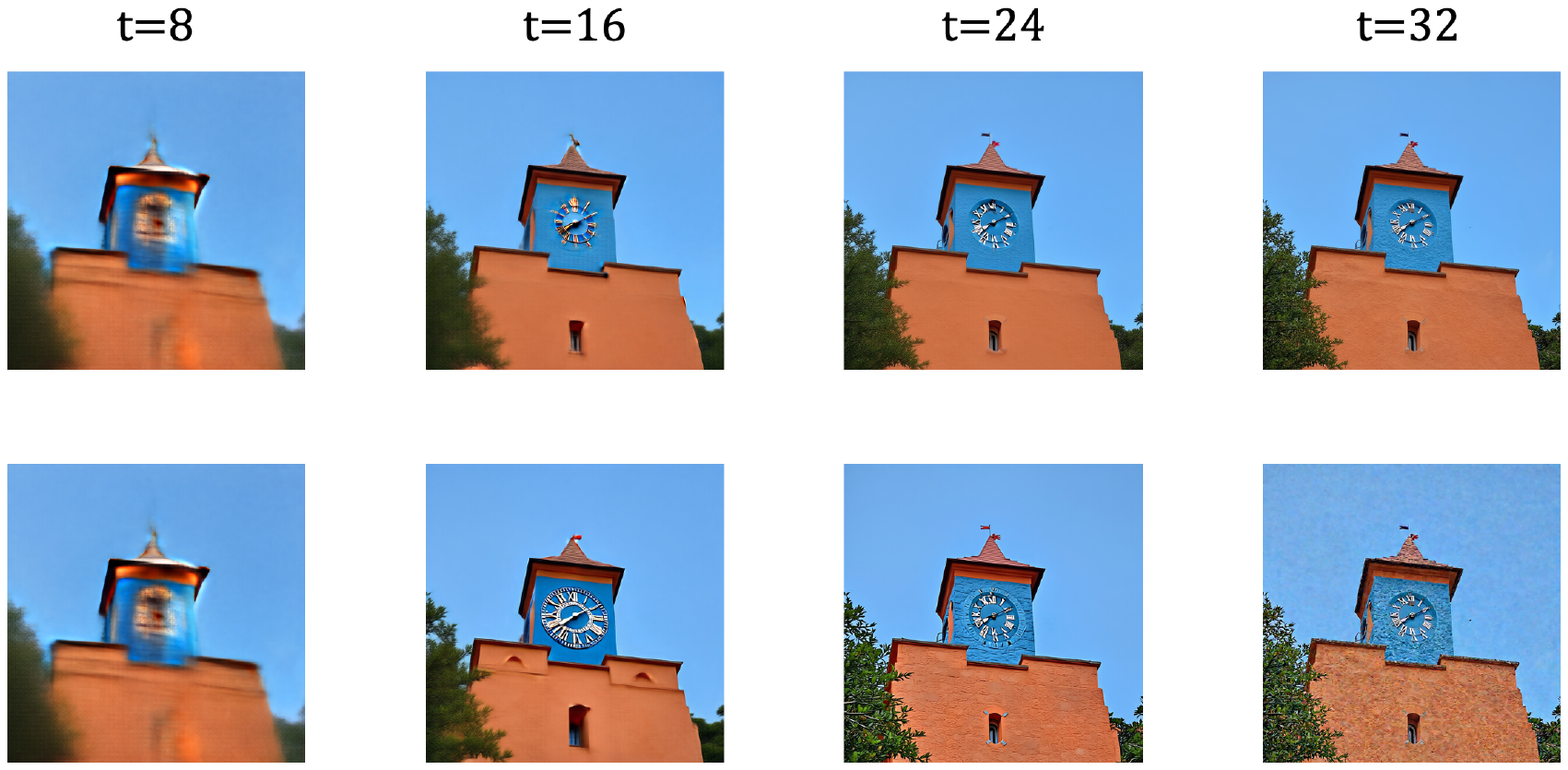} 
    \vspace{-40pt}
    \caption{
        Trajectory Divergence between Authentic and Piecewise Paths.
        (\textit{Top}) States from the teacher's authentic, continuous 32-step trajectory at key timesteps ($t=8, 16, 24, 32$).
        (\textit{Bottom}) States from a piecewise trajectory, where each 8-step stage is re-initialized via linear interpolation. A visible divergence emerges at $t=16$ and grows.
    }
    \label{trajectory_divergence} 
\end{figure}

\noindent\paragraph{Online Trajectory Alignment} 

We identify that PeRFlow's performance is fundamentally limited by a critical training-inference mismatch problem that affects both the teacher and the student, the two most important components in distillation.
For the teacher, this leads to a \textit{Teacher Trajectory Mismatch}, as shown in Figure~\ref{fig:mismatch_a} and Figure~\ref{trajectory_divergence}. During training, the teacher denoises from linearly interpolated starting points that lies off its authentic inference trajectory. Consequently, the distillation targets provided to the student are based on suboptimal trajectories that the teacher would never follow during actual inference, fundamentally damages the guidance.
For the student, this causes an \textit{Inter-Stage Distribution Mismatch}, as illustrated in Figure~\ref{fig:mismatch_b}. The training process initializes each stage with a fresh interpolated start point; while during inference, each stage receives the output from the previous stage. In Section~\ref{limitations in perflow}, we prove that this mismatch is inherent to PeRFlow's design, persisting even under a perfect optimization. 
Our Online Trajectory Alignment (OTA) resolves both problems by training on authentic teacher trajectories, ensuring the teacher operates on-trajectory and training matches inference 
distributions.

\vspace{-0.5cm}
\noindent\paragraph{Adversarial distillation on trajectory} 
Adversarial distillation has been proved as a well-established technique for enhancing few-step distillation~\cite{sdxllightning, ufogen, apt, add, ladd, flash, dmd2}, but its application on ReFlow ramains unexplored. We propose to directly impose an adversarial loss between the teacher and student trajectories. This encourages the student trajectory to more faithfully mimic the teacher's, ultimately resulting in significant performance gains.
Furthermore, we identify and rectify a critical flaw in the standard \texttt{FlowMatchEulerDiscreteScheduler} that impairs performance in the few-step generation. 

\vspace{0.1cm}
In summary, our contributions are as follows.
\begin{itemize}
    \item We resolve a fundamental issue in PeRFlow. We reveals that PeRFlow suffers from a training-inference distribution mismatch, causing suboptimal teacher trajectory and student's error. We propose OTA to solve it.

    \item We introduce adversarial distillation to ReFlow-based method. By applying it on ODE trajectory, discriminator forces the student's trajectory to mimic the teacher's.

    \item We find and fix a previously overlooked flaw in the widely used \texttt{FlowMatchEulerDiscreteScheduler} that harms few-step generation. By fixing it with only a few lines of code, we demonstrate our improved scheduler boosts performance for both few-step distillation and distillation-free pretrained models.

\end{itemize}
\section{Related Work}
\label{sec:related work}

\vspace{-0.65cm}
\noindent\paragraph{Diffusion Model}
Diffusion models~\cite{ediff, ddpm, edm, songscore} have established themselves as a leading class of generative models, renowned for their ability to synthesize high-quality and diverse samples. The core idea involves a two-stage process: a forward process that gradually adds noise over many steps, and a reverse process where a neural network is trained to denoise the smaple step-by-step~\cite{2015deep, ddpm}. Within inference, the model starts from pure noise and iteratively applies this learned denoising function, progressively refining the noise into a coherent sample. While it achieves great success in visual generation~\cite{qwenimage, sd3, stepvideo, yinvideo}, audio generation~\cite{makeaudio, tian2025audiox} and text generation~\cite{lldm, dream7b}, a major limitation remains. The iterative nature of the reverse process requires numerous sequential steps~\cite{ddim}, making inference slow. Therefore, improving the inference speed of diffusion models has become a key area of research.

\vspace{-0.3cm}
\noindent\paragraph{Diffusion Acceleration}
Among acceleration methods, step distillation has emerged as a prominent strategy . Its basic goal is to train a student model to replicate the output of a multi-step teacher in fewer steps and can be categorized into Progressive Distillation(PD)~\cite{pd}, Consistency Distillation(CD)~\cite{cm, lcm, pcm, hypersd, icm, simplifyingcm, ctm}, Distribution Matching Distillation(DMD)~\cite{dmd, dmd2}, Adversarial Aistillation~\cite{add, ladd, ufogen, apt} and Rectified Flow(ReFlow)~\cite{flowmatch, instaflow, perflow, rectifieddiffusion}. 
To mitigate the high computational cost of this approach, PD introduced a staged process that iteratively halves the number of sampling steps, though this multi-stage training can be cumbersome. 
CD~\cite{cm, lcm, icm, simplifyingcm} trains the student to map any point along a single generation trajectory to the same final output. This concept was later relaxed into trajectory consistency~\cite{ctm, pcm}. 
Rather than enforcing point-wise correspondence along a trajectory, DMD~\cite{dmd, dmd2} focus on matching the student's output distribution to the real distribution using score function theory~\cite{songscore}. 
To further improve the perceptual quality of the distilled models, adversarial distillation~\cite{add} employs a discriminator to align the student's output distribution with a target, which can be the real data distribution~\cite{ufogen} or output distribution from the teacher model~\cite{ladd}.
Besides, ReFlow~\cite{instaflow} aims to straighten the generation trajectory into a linear path for efficient sampling. This concept was advanced by PerFlow~\cite{perflow}, which uses a piecewise linear trajectory to improve few-step results. However, their performance has struggled to match other top-tier methods in previous work~\cite{hypersd, shoutcut}, hindering their practical application. 

\vspace{-0.3cm}
\noindent\paragraph{Flow Matching and Rectified Flow} ReFlow is rooted in Flow Matching~\cite{flowmatch, flowstraight}, which models the generation process as a continuous flow from a simple noise distribution to the data distribution by learning a corresponding velocity field. Sampling is then performed by solving an ordinary differential equation (ODE) guided by this learned field. However, a key challenge is that noise-data pairs used in training are unorganized, causing the learned marginal velocity field is typically curved~\cite{minimizing}, leading to significant errors when using large step sizes in numerical ODE solvers. To address this, Rectified Flow~\cite{instaflow, rectifieddiffusion, improving} introduces an iterative "reflow" procedure. It repeatedly generates data pairs from a trained model to enable organized noise-image pairs and uses them to train a subsequent model, progressively creating straighter and more direct trajectories that are easier to solve with fewer steps. This core idea has been extended in various ways; for instance, InstaFlow~\cite{instaflow} applied consecutive reflow processes to large-scale text-to-image models, while PeRFlow~\cite{perflow} improved few-step performance by partitioning the trajectory and applying localized straightening within each segment.
\section{Background on PeRFlow}
\label{sec:preliminary}

Piecewise Rectified Flow (PeRFlow) was proposed to improve reflow in few-step generation. It employs a divide-and-conquer strategy, partitioning the full generation trajectory from time $t=1$ to $t=0$ into $K$ distinct stages, $\{[t_k, t_{k-1})\}_{k=1}^K$.

The core idea is to approximate the teacher's curved trajectory within each stage as a straight line. To achieve this, PeRFlow generates training pairs $(z_{t_k}, z_{t_{k-1}})$ for each stage. The starting point $z_{t_k}$ is computed by adding noise to a real image $z_0$ according to the teacher's forward process:
\begin{equation}
    z_{t_k} = \sigma_k \epsilon + (1 - \sigma_k) z_0,
    \label{eq:perflow_start}
\end{equation}
where $\sigma_k = \sigma(t_k)$ and $\epsilon \sim \mathcal{N}(0, I)$. The corresponding endpoint $z_{t_{k-1}}$ is then obtained by using the teacher model as an ODE solver $\Phi$ to evolve $z_{t_k}$ to time $t_{k-1}$, i.e., $z_{t_{k-1}} = \Phi(z_{t_k}, t_k, t_{k-1})$.

The student model, parameterized as a velocity field $v_\theta(z_t, t)$, is trained to match the constant velocity required to travel from $z_{t_k}$ to $z_{t_{k-1}}$. The target velocity is defined as $v^* = (z_{t_{k-1}} - z_{t_k}) / (t_{k-1} - t_k)$. The model is optimized by minimizing the following loss over all stages:
\begin{equation}
    \mathcal{L} = \sum_{k=1}^{K} \mathbb{E} \int_{t_{k-1}}^{t_k} \| v_\theta(z_t, t) - v^* \|^2 dt,
    \label{eq:perflow_loss}
\end{equation}
where $z_t$ is a linear interpolation between $z_{t_k}$ and $z_{t_{k-1}}$. 

\section{Method}

\subsection{Limitations in PeRFlow}
\label{limitations in perflow}

\begin{table}[t!] 
    \vspace{10pt}
    \small 
    \centering
    \captionof{table}{
        We sample 1k prompts from COCO. Starting from an interpolated point significantly degrades teacher performance.
    }
    \label{table:trajectory_mismatch}
    \begin{tabular}{@{}lrr@{}}
        \toprule
        \textbf{Method} & \textbf{PickScore} $\uparrow$ & \textbf{HPSv2} $\uparrow$ \\
        \midrule
        Teacher (Authentic) & 22.58 & 29.43 \\
        Teacher (Interpolated) & 22.39 & 28.63 \\
        \bottomrule
    \end{tabular}
\end{table}

\begin{table*}[t!]
\centering
\footnotesize
\begin{minipage}[t]{0.48\textwidth}
    \begin{algorithm}[H]
    \caption{PeRFlow Training}
    \label{alg:perflow}
    \small
    \textbf{Input:} Teacher $v_T$, Student $v_S$, Dataset $\mathcal{D}$, Stages $K$
    \begin{algorithmic}[1]
    \State Initialize student model $v_S$
    \Repeat
        \State Sample $z_0 \sim \mathcal{D}$ and $\epsilon \sim \mathcal{N}(0, I)$
        \State Sample stage $k \in \{1, \dots, K\}$
        \State \textbf{\textcolor{red}{
               $z_{t_k} \leftarrow (1-t_k)z_0 + t_k\epsilon$}} 
               \hfill \textcolor{gray}{\small\textit{// Off-Trajectory}}
        \State $z_{t_{k-1}} \leftarrow \text{ODESolve}(v_T, z_{t_k}, t_k, t_{k-1})$
        \State $v_{\text{target}} \leftarrow (z_{t_{k-1}} - z_{t_k}) / (t_{k-1} - t_k)$
        \State Sample $t \sim U(t_{k-1}, t_k)$
        \State $z_t \leftarrow \text{Interpolate}(z_{t_k}, z_{t_{k-1}}, t)$
        \State $\mathcal{L} \leftarrow \| v_S(z_t, t) - v_{\text{target}} \|^2$
        \State Update $v_S$ using $\nabla \mathcal{L}$
    \Until{convergence}
    \end{algorithmic}
    \end{algorithm}
\end{minipage}%
\hfill
\begin{minipage}[t]{0.48\textwidth}
    \begin{algorithm}[H]
    \caption{OTA Training (Ours)}
    \label{alg:OTA}
    \small
    \textbf{Input:} Teacher $v_T$, Student $v_S$, Dataset $\mathcal{D}$, Stages $K$
    \begin{algorithmic}[1]
    \State Initialize student model $v_S$
    \Repeat
        \State Sample $z_0 \sim \mathcal{D}$ and $\epsilon \sim \mathcal{N}(0, I)$
        \State Sample stage $k \in \{1, \dots, K\}$
        \State \textbf{\textcolor{green!60!black}{
               $z_{t_k} \leftarrow \text{ODESolve}(v_T, \epsilon, 1, t_k)$}} 
               \hfill \textcolor{gray}{\small\textit{// On-Trajectory}}
        \State $z_{t_{k-1}} \leftarrow \text{ODESolve}(v_T, z_{t_k}, t_k, t_{k-1})$
        \State $v_{\text{target}} \leftarrow (z_{t_{k-1}} - z_{t_k}) / (t_{k-1} - t_k)$
        \State Sample $t \sim U(t_{k-1}, t_k)$
        \State $z_t \leftarrow \text{Interpolate}(z_{t_k}, z_{t_{k-1}}, t)$
        \State $\mathcal{L} \leftarrow \| v_S(z_t, t) - v_{\text{target}} \|^2$
        \State Update $v_S$ using $\nabla \mathcal{L}$
    \Until{convergence}
    \end{algorithmic}
    \end{algorithm}
\end{minipage}
\end{table*}

The piecewise training strategy of PeRFlow introduces two fundamental limitations, and lead to error accumulation during inference.

\noindent\paragraph{Teacher Trajectory Mismatch}

The first limitation arises from the definition of the initial conditions for each segmented trajectory. The protocol prescribes the starting point of each stage as a linear interpolation between the final image $z_0$ and initial noise $z_1$. This formulation differs from a real inference process, where the state at time $t_k$ is the result of evolving from an earlier state along the teacher's non-linear vector field. Crucially, the point derived from this prescribed interpolation does not coincide with the corresponding state on the teacher's true trajectory.

We provide empirical evidence for this performance gap in Figure~\ref{trajectory_divergence} (visual) and Table~\ref{table:trajectory_mismatch} (quantitative). 
Figure~\ref{trajectory_divergence} meticulously illustrates how the piecewise path diverges from the authentic one. The \textit{top row} displays intermediate states ($t=8, 16, 24, 32$) from a single, continuous 32-step inference by the teacher model. The \textit{bottom row} simulates the PeRFlow data-making process: the state at $t=8$ is generated by running the teacher for 8 steps from pure noise, which aligns perfectly with the authentic path. However, the state at $t=16$ is generated by starting from a linearly interpolated point at $t=8$ and then running the teacher for another 8 steps. This re-initialization immediately forces the model onto a suboptimal, \textbf{``off-trajectory''} path, causing a visible and growing discrepancy at $t=16$ and beyond.
Since the states shown in the figure are intermediate steps and thus not suitable for final image quality assessment, we designed a targeted experiment to quantify the performance degradation on the final output. As detailed in Table~\ref{table:trajectory_mismatch}, we compare two settings on 1000 samples from the COCO dataset: (1) the teacher's standard, uninterrupted 32-step inference, which serves as the true performance benchmark; and (2) a process mimicking the final segmented trajectory, where we start from a linear interpolation at $t=24$ and run the teacher for the remaining 8 steps. The resulting drop in PickScore and HPSv2 confirms that this misalignment of stage boundaries imposes a fundamental and measurable limit on the achievable performance.

\noindent\paragraph{Inter-Stage Distribution Mismatch}
Another critical limitation is a distribution mismatch at the boundaries between consecutive stages. During training, the starting point of each stage, $z_{t_k}$, is defined as a linear interpolation between a ground-truth image $z_0$ and pure noise. However, during inference, the process is sequential: the output of the first stage serves as the input for the second. We formally prove that these two distributions—the ideal starting point for the next stage and the actual output from the current stage—are inherently different. This mismatch is unavoidable unless the teacher model is already a perfect Rectified Flow model, even if the student is perfectly trained. The detailed proof is provided in Supplementary Material~\ref{sec:appendix_mismatch_proof}.

\begin{table*}[t!]
\centering
\caption{Performance comparison between the original and our improved scheduler under few-step inference. Our method demonstrates superior performance, especially at 4 steps, which is critical for distilled models. Experiments are conducted on SD3M.}
\label{tab:scheduler_comparison}
\begin{tabular}{lcccc}
\toprule
\multirow{2}{*}{\textbf{Setting}} & \multicolumn{2}{c}{\textbf{Original Scheduler}} & \multicolumn{2}{c}{\textbf{Our Improved Scheduler}} \\
\cmidrule(lr){2-3} \cmidrule(l){4-5}
 & \textbf{PickScore $\uparrow$} & \textbf{HPSv2 $\uparrow$} & \textbf{PickScore $\uparrow$} & \textbf{HPSv2 $\uparrow$} \\
\midrule
Shift = 3, N = 32 & 22.61 & 29.50 & 22.62 & 29.51 \\
Shift = 3, N = 10 & 21.86 & 26.55 & 21.95 & 27.04 \\
Shift = 3, N = 4 & 19.31 & 15.91 & 20.11 & 19.34 \\
\midrule
Shift = 1, N = 32 & 22.27 & 28.04 & 22.29 & 28.10 \\
Shift = 1, N = 10 & 20.56 & 21.09 & 20.74 & 21.88 \\
Shift = 1, N = 4 & 18.50 & 12.11 & 18.97 & 14.15 \\
\bottomrule
\end{tabular}
\end{table*}

\subsection{Online Trajectory Alignment}

Motivated by \textit{backward distillation}~\cite{imagineflash}, we introduce Online Trajectory Alignment (OTA) to rectify the deficiencies inherent in PeRFlow's piecewise strategy. Specifically, the starting state $z_{t_k}$ for a given stage is no longer derived from a static interpolation.

As delineated in Algorithm~\ref{alg:OTA}, the starting state $z_{t_k}$ for a given stage $[t_k, t_{k-1}]$ is no longer derived from a linear interpolation between data and noise. Instead, it is generated by solving the teacher's probability flow ODE from initial noise $z_1 \sim \mathcal{N}(0, I)$ down to time $t_k$:
\begin{equation}
    z_{t_k} = z_1 + \int_{1}^{t_k} v_T(z(s), s) ds.
\end{equation}
This on-policy state generation mechanism yields two principal advantages that directly counteract the limitations of PeRFlow. First, $z_{t_k}$ resides on the teacher's true trajectory resolves the \textit{teacher trajectory mismatch} and allowing the student to learn from the teacher's uncompromised dynamics. Second, this approach ensures that the distribution of starting points in training mirrors the distribution of intermediate states encountered during sequential inference, thus closing the \textit{inter-stage distribution mismatch}.
The algorithmic distinction is detailed in Table~\ref{alg:OTA}. Although OTA incurs an additional computational cost for the online generation of $z_{t_k}$, this is a principled trade-off that proves essential for minimizing error propagation and achieving a higher-fidelity distillation.

\subsection{Adversarial Distillation on ODE Trajectory}

To further enhance the fidelity of the few-step student model, we integrate an adversarial training objective into our framework. This encourages the student to produce trajectories that are perceptually indistinguishable from the teacher's. Our discriminator utilizes the backbone of a pre-trained diffusion model, and following LADD~\cite{ladd}, we append a discriminator head to it.

Within our experiments, we focus on 4-step distillation. In this context, the student model ($v_S$) acts as a generator, and its goal is to produce a 4-step trajectory to mimic teacher model($v_T$)'s many-steps trajectory.
The training is guided by two main loss components. The first is the adversarial loss, $\mathcal{L}_{\text{adv}}$. The student operates over a discrete set of timesteps, $\mathcal{T}_S = \{t_1, t_2, t_3, t_4\}$. We sample a timestep $t$ from this set using a probability distribution $p(t)$. Let $z_t^T$ be a state on the teacher's trajectory and $z_t^S$ be the corresponding state on the student's. The student is trained to maximize the discriminator's output for its generated states:
\begin{equation}
    \mathcal{L}_{\text{adv}} = - \mathbb{E}_{z_t^S \sim v_S, t \sim p(t)} [D(z_t^S, t)].
\end{equation}

To stabilize training, we incorporate a feature matching loss~\cite{dcm}. This loss minimizes the L2 distance between intermediate feature representations from the student and teacher within the discriminator. Let $D_l(z_t, t)$ be the feature map from the $l$-th layer of the discriminator's backbone. The loss is defined as:
\begin{equation}
    \mathcal{L}_{\text{FM}} = \sum_{l=1}^{L} \mathbb{E}_{z_t^S, z_t^T, t \sim p(t)} \left[ \| D_l(z_t^T, t) - D_l(z_t^S, t) \|_2 \right],
\end{equation}
where $L$ is the number of backbone layers. The student's final objective is a weighted sum of the distillation loss, the adversarial loss, and the feature matching loss:
\begin{equation}
    \mathcal{L}_{\text{student}} = \mathcal{L}_{\text{dist}} + \lambda_{\text{adv}}\mathcal{L}_{\text{adv}} + \lambda_{\text{FM}}\mathcal{L}_{\text{FM}}.
\end{equation}

\begin{figure*}[t]
    \vspace{-80pt}
    \noindent
    \includegraphics[width=\textwidth]{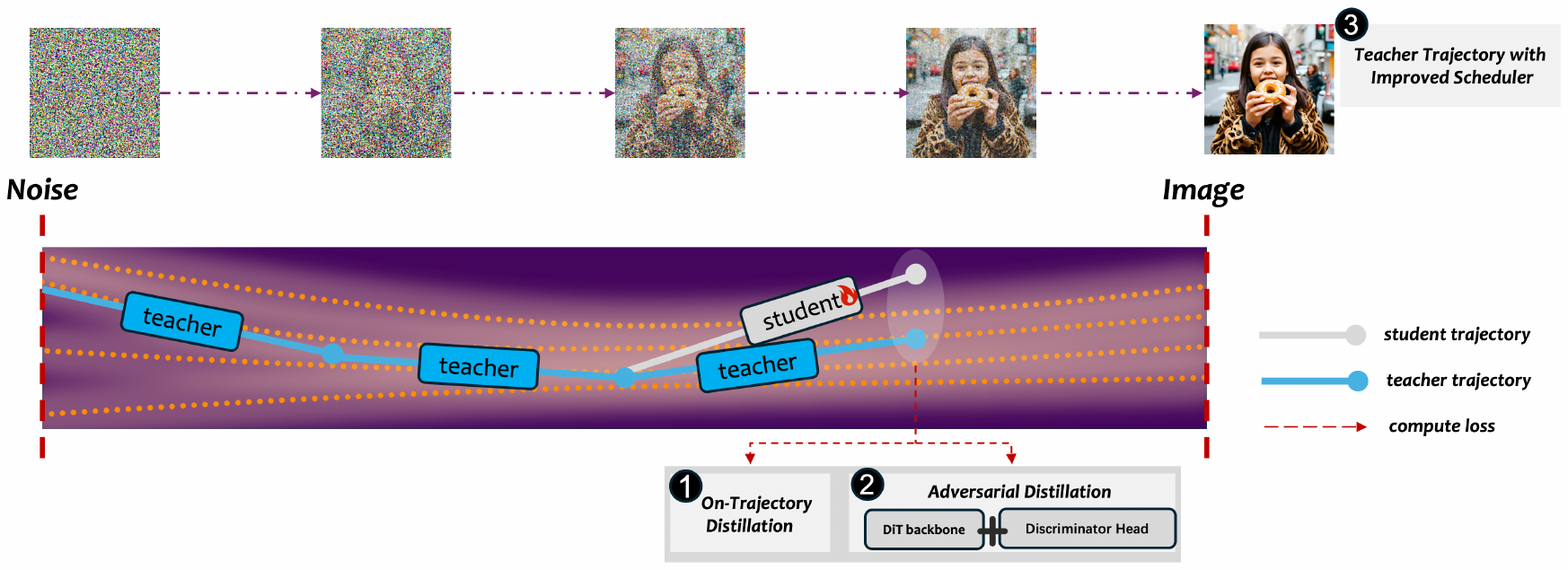} 
    \vspace{-100pt}
    \caption{An overview of our proposed FlowSteer. (1) OTA online generates on-trajectory starting points for each stage by simulating the teacher's own process. (2) A discriminator combined by DiT backbone and discriminator head force the student's trajectory to mimic teacher's. (3) We build sub-trajectory basicly on our improved scheduler, for both teacher and student.}
    \label{main_figure}
\end{figure*}

\newcommand{\stackheader}[2]{\begin{tabular}{@{}c@{}}\textbf{#1}\\\textbf{#2}\end{tabular}}

\begin{table*}[ht] 
\centering 

\setlength{\tabcolsep}{4pt}

\caption{\textbf{Quantitative comparison on the COCO 10k and GenEval.} 
NFE denotes the Number of Function Evaluations. 
$^\dagger$ indicates our implementation.
Obj.: Object;
Attr.: Attribution Binding.
}
\vspace{-2mm}
\label{tab:main_results}

\resizebox{\linewidth}{!}{
\begin{tabular}{@{\extracolsep{\fill}}lccccccccccc}
\toprule
\multirow{2}{*}{\textbf{Method}} & \multirow{2}{*}{\textbf{NFE}} & \multicolumn{3}{c}{\textbf{COCO 10k}} & \multicolumn{7}{c}{\textbf{GenEval}} \\
\cmidrule(lr){3-5} \cmidrule(lr){6-12}
& & \textbf{PickScore} & \textbf{HPSv2} & \textbf{CLIP Score} & \textbf{Overall} & \textbf{Single Obj.} & \textbf{Two Obj.} & \textbf{Counting} & \textbf{Colors} & \textbf{Position} & \textbf{Attr.} \\
\midrule
\multicolumn{12}{@{}l}{\textit{Backbone: SD3-Medium}} \\
\midrule

\textcolor{gray}{SD3-Medium (Pre-trained)} & \textcolor{gray}{$20 \times 2$} & \textcolor{gray}{22.41} & \textcolor{gray}{28.02} & \textcolor{gray}{32.78} & \textcolor{gray}{0.6639} & \textcolor{gray}{0.9875} & \textcolor{gray}{0.8131} & \textcolor{gray}{0.5281} & \textcolor{gray}{0.8298} & \textcolor{gray}{0.2625} & \textcolor{gray}{0.5625} \\
PCM~\cite{pcm} (Shift=1) & $4 \times 2$ & 22.12 & 27.81 & 32.32 & 0.6121 & 0.9906 & 0.6919 & 0.5375 & 0.8298 & 0.2125 & 0.4100 \\
PCM~\cite{pcm} (Shift=3) & $4 \times 2$ & 22.28 & 27.68 & 32.05 & 0.6339 & 0.9969 & 0.7273 & 0.5656 & 0.8059 & 0.2175 & 0.4900 \\
PCM~\cite{pcm} (Stochastic) & $4 \times 2$ & 22.07  & 27.79 & 32.50 & 0.6247 & 0.9906 & 0.7576 & 0.5344 & 0.8059 & 0.2250 & 0.4350 \\
Hyper-SD~\cite{hypersd} & $4 \times 2$ & 22.28 & 28.04 & 32.31 & 0.6336 & 0.9969 & 0.7601 & 0.5594 & 0.8324 & 0.2000 & 0.4525 \\
Flash Diffusion~\cite{flash} & $4 \times 1$ & 22.37 & 27.35 & 32.51 & 0.6672 & 0.9938 & 0.7551 & 0.5844 & \textbf{0.8750} & 0.2950 & 0.5000 \\
PeRFlow$^\dagger$~\cite{perflow} & $4 \times 1$ & 22.19 & 26.36 &  32.55 & 0.6357 & 0.9969 & 0.7289 & 0.5625 & 0.7952 & 0.2650 & 0.4650 \\ 
\rowcolor{gray!20}
\textbf{Ours} & \textbf{$4 \times 1$} & \textbf{22.39} & \textbf{28.60} & \textbf{32.81} & \textbf{0.6859} & \textbf{1.0000} & \textbf{0.8182} & \textbf{0.6125} & 0.8245 & \textbf{0.3150} & \textbf{0.5450} \\
\midrule
\multicolumn{12}{@{}l}{\textit{Backbone: SD3.5-Large}} \\
\midrule
PeRFlow$^\dagger$~\cite{perflow} & $4 \times 1$ & 22.29 & 27.14 & 32.73 & 0.6439 & 0.9875 & 0.7778 & 0.5906 & 0.7952 & 0.2735 & 0.4750 \\ 
\rowcolor{gray!20}
\textbf{Ours} & \textbf{$4 \times 1$} & \textbf{22.61} & \textbf{28.47} & \textbf{32.87} & \textbf{0.6780} & \textbf{0.9906} & \textbf{0.8561} & \textbf{0.6406} & \textbf{0.8005} & \textbf{0.2475} & \textbf{0.5325} \\
\bottomrule
\end{tabular}
}
\end{table*}

\subsection{FlowMatchEulerDiscreteScheduler is flawed}

The standard \texttt{FlowMatchEulerDiscreteScheduler}, while effective for a large number of inference steps, exhibits a structural flaw in the few-step regime. The process begins by setting up a full schedule of 1000 timesteps. For an $N$-step inference, the original scheduler samples $N$ points from this list and then separately appends the final state where $\sigma=0$. This leads to a critical issue: the step size from the last sampled sigma to zero is not proportional to the other steps. For instance, with a shift of 3, this jump is from a sigma of 0.0089 to 0, which can degrade quality in few-step generation.
To address this, we propose an improved scheduler. After setting the initial 1000 timesteps, we first augment the schedule by appending the terminal state ($\sigma=0$). Then, for an $N$-step inference, we sample $N+1$ points using a linear space across this complete, augmented range. This ensures that all step intervals, including the final one to zero, are proportionally scaled. The pseudo-code for this improved scheduler is provided in Supplementary Material~\ref{sec:appendix_scheduler}, and its performance comparison is detailed in Table~\ref{tab:scheduler_comparison}.

The results in Table~\ref{tab:scheduler_comparison} demonstrate, the performance difference between the two schedulers is negligible at a high number of inference steps. However, as the number of steps decreases, our improved scheduler shows a clear and significant advantage over the original one in both PickScore and HPSv2 metrics. It implies our improved scheduler is more suitable for step distillation, where inference is almost exclusively performed in the few-steps condition.

\section{Experiment}

\begin{figure*}[t]
    \centering
    \includegraphics[trim=0pt 300pt 0pt 300pt, clip, width=\linewidth]{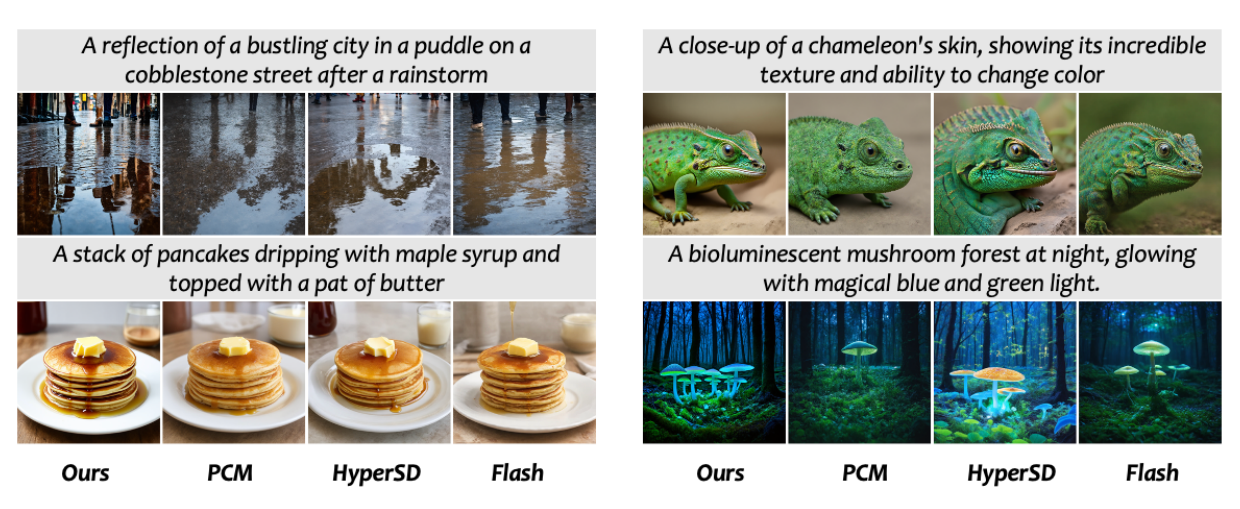}
    \vspace{-15pt}
    \caption{Qualitative comparison with other methods at 4 steps(8 NFE for PCM and 4 NFE for others). Our model generates images with higher quality and better text-image consistency.}
    \label{fig:qualitative_comparison}
\end{figure*}

\begin{figure*}[t]
    \centering
    \includegraphics[trim=50pt 300pt 50pt 300pt, clip, width=\linewidth]{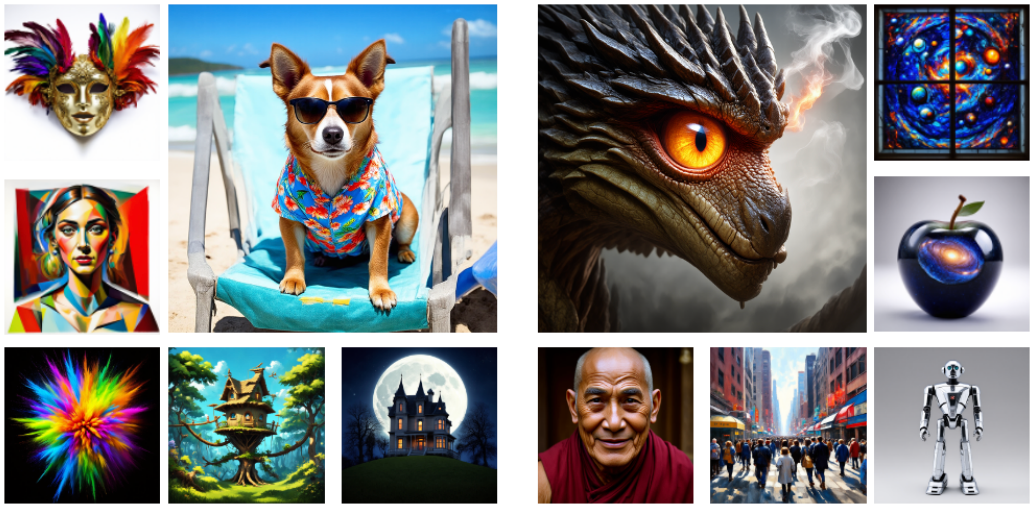}
    \vspace{-15pt}
    \caption{Diverse high-quality images generated by our method using four NFE. These examples demonstrate our proposed method can produce aesthetically appealing and diverse results.}
    \label{fig:qualitative_showcase}
\end{figure*}

In this section, we conduct experiments to validate the effectiveness of our proposed method. Our approach integrates several key components, including online trajectory alignment, adversarial distillation and improved scheduler. We first outline the experimental setup, then present quantitative and qualitative results, and conclude with a detailed ablation study.

\subsection{Experiment Configuration}

\paragraph{Models and Datasets}
Our experimental is based on two advanced text-to-image diffusion models: Stable Diffusion 3 Medium and Stable Diffusion 3.5 Large. Both models leverage the Multimodal Diffusion Transformer (MMDiT)~\cite{sd3} architecture. We use the prompt of Flux-Reason-6M~\cite{fluxreason6m} dataset as our training dataset and creates generation trajectories online. To enable efficient adaptation, we employ LoRA~\cite{lora} training for both two, significantly reducing the number of trainable parameters while preserving the integrity of the original pretrained weights.

\vspace{-10pt}
\paragraph{CFG Distillation}
We aim to create a guidance-free generator to accelerate the inference process. To this end, we employ the CFG-Distillation methodology~\cite{cfgdistillation}, which transfers the knowledge from a guided teacher model to a guidance-free student model. The teacher model using a Classifier-Free Guidance scale sampled uniformly from the interval $\mathcal{U}[7, 13]$  and the student model is trained on a fixed guidance scale at $\omega=0$. This process bakes the text-guidance into the model's weights, eliminating the need for the dual forward pass required by CFG at inference time.

\subsection{Quantitative Results}
The main quantitative results are presented in Table~\ref{tab:main_results}. 
On the SD3-Medium backbone, our method demonstrates a clear performance advantage over existing methods on NFE of 4. Firstly,  our approach significantly surpasses PeRFlow~\cite{perflow}, also evidenced by the results on the SD3.5-Large.
We also observe that our method outperforms CD-based methods(PCM~\cite{pcm} and Hyper-SD~\cite{hypersd}) and DMD-based method(Flash Diffusion~\cite{flash}). It is worth noting that both PCM and Hyper-SD utilize classifier-free guidance (CFG) with scales of 1.2 and 3, respectively. This necessitates two forward passes for both methods, effectively doubling their inference time compared to other methods. Since disabling this guidance results in a significant drop in image quality, we followed their official settings to ensure a fair comparison.

\subsection{Qualitative Comparison}

In addition to quantitative analysis, we provide a visual evaluation in Figure~\ref{fig:qualitative_comparison} and Figure~\ref{fig:qualitative_showcase}. Specifically, Figure~\ref{fig:qualitative_comparison} showcases a direct comparison between our method and other methods using 4 sampling steps. Figure~\ref{fig:qualitative_showcase} presents a collection of high-quality images generated by our model. Both of them demonstrate our model can produce high-quality and aestheticly appeaing images.

\subsection{Ablation Study}

To validate our design choices and understand the contribution of each component, we conduct a series of comprehensive ablation studies on the SD3-Medium model. We structure our analysis into three parts: we first determine the optimal discriminator configuration, then analyze the timestep sampling strategy, and finally evaluate the impact of our core methodological contributions. All experiments are conducted on SD3M using 4-step inference.

\subsubsection{Discriminator Configuration}
We begin by investigating the optimal architectural and training settings for the discriminator. We ablate four key aspects: the backbone depth, the number of discriminator heads, the backbone training strategy, and the GAN loss function. The results are presented in Table~\ref{tab:discriminator_ablation}. We find a 12-block backbone provides the best trade-off between model capacity and performance. A single discriminator head on the final block's features, which provides a global assessment, is more effective than per-block supervision. Full fine-tuning of the backbone is crucial, significantly outperforming parameter-efficient methods. Finally, consistent with modern GAN training practices, Hinge Loss delivers the most stable and effective results.

\begin{table*}[ht!]
\small
\caption{\textbf{Ablation study on the discriminator configuration.} The best-performing setting in each category is highlighted. We choose some blocks of pretrained diffusion model as backbone and then add some discriminator heads to it.}
\label{tab:discriminator_ablation}
\begin{minipage}[t]{0.48\textwidth}
    \centering
    \textbf{(a) Effect of backbone depth}
    \vspace{0.3em}
    \vspace{0pt}
    \begin{tabular}{@{} l cc @{}} 
    \toprule
    Backbone Depth & PickScore $\uparrow$ & HPSv2 $\uparrow$ \\
    \midrule
    6 Blocks & 22.13 & 27.55 \\
    \textbf{12 Blocks} & \textbf{22.39} & \textbf{28.60} \\
    18 Blocks & 22.16 & 27.80 \\
    24 Blocks (Full) & 22.29 & 28.25 \\
    \bottomrule
    \end{tabular}
    
    \vspace{1em} 
    \textbf{(c) Effect of discriminator heads}
    \vspace{0.3em}
    
    \begin{tabular}{@{} l cc @{}} 
    \toprule
    Discriminator Heads & PickScore $\uparrow$ & HPSv2 $\uparrow$ \\
    \midrule
    \textbf{1 Head (on last block)} & \textbf{22.39} & \textbf{28.60} \\
    12 Heads (Per-block) & 22.15 & 27.44 \\
    \bottomrule
    \end{tabular}
\end{minipage}
\centering
\hfill
\begin{minipage}[t]{0.48\textwidth}
    \centering
    \vspace{0pt}
    
    \textbf{(b) Effect of backbone training strategy}
    \vspace{0.3em}
    \begin{tabular}{@{} l cc @{}} 
    \toprule
    Backbone Training & PickScore $\uparrow$ & HPSv2 $\uparrow$ \\
    \midrule
    Frozen & 22.28 & 27.45 \\
    LoRA & 22.27 & 28.07 \\
    \textbf{Full Fine-tuning} & \textbf{22.39} & \textbf{28.60} \\
    \bottomrule
    \end{tabular}
    
    \vspace{1em}
    \textbf{(d) Effect of GAN loss function}
    \vspace{0.3em}
    
    \begin{tabular}{@{} l cc @{}} 
    \toprule
    GAN Loss Function & PickScore $\uparrow$ & HPSv2 $\uparrow$ \\
    \midrule
    LSGAN & 22.32 & 27.64 \\
    WGAN & 21.96 & 27.28 \\
    \textbf{Hinge Loss} & \textbf{22.39} & \textbf{28.60} \\
    \bottomrule
    \end{tabular}
\end{minipage}
\end{table*}

\begin{table*}[t]
\small
\begin{minipage}[t]{0.48\textwidth}
\centering
    \caption{\textbf{Ablation on timestep sampling probabilities over four discrete generation stages.} The chosen strategy is highlighted.}
    \label{tab:timestep_ablation}
    \begin{tabular}{@{} l cc @{}}
        \toprule
        \textbf{Timesteps' probabilities} & \textbf{PickScore $\uparrow$} & \textbf{HPSv2 $\uparrow$} \\
        \midrule
        \textbf{0.4, 0.2, 0.2, 0.2} & \textbf{22.39} & 28.60 \\
        0.4, 0.1, 0.1, 0.4 & 22.23 & \textbf{28.69} \\
        0.3, 0.2, 0.2, 0.3 & 22.27 & 28.50 \\
        0.4, 0.3, 0.2, 0.1 & 22.17 & 27.98 \\
        0.1, 0.2, 0.3, 0.4 & 22.31 & 28.35 \\
        \bottomrule
    \end{tabular}
\end{minipage}
\hfill 
\begin{minipage}[t]{0.48\textwidth}
\centering
\caption{\textbf{Ablation study on the key components of our method.}}
\label{tab:component_ablation}
    \begin{tabular}{@{} l cc @{}}
    \toprule
    \textbf{Method Configuration} & \textbf{PickScore} $\uparrow$ & \textbf{HPSv2} $\uparrow$ \\
    \midrule
    Baseline & 22.19 & 26.36 \\
    \midrule
    Baseline + OTA & 22.23 & 26.79 \\
    Baseline + Adv. Distillation & 22.18 & 27.73 \\
    Baseline + Scheduler & \textbf{22.50} & 27.75 \\
    \midrule
    \textbf{Ours (Full Method)} & 22.39 & \textbf{28.60} \\
    \bottomrule
    \end{tabular}
\end{minipage}
\end{table*}

\subsubsection{Timestep Sampling Strategy}
The allocation of sampling steps across the diffusion trajectory is pivotal for optimizing few-step generation. In our 4-step framework, we explore how different probability distributions for timestep $T$ sampling impact final image quality. We conducted an ablation study by partitioning the generation process into four stages (from high to low noise) and assessing five distinct distributions as shown in Table~\ref{tab:timestep_ablation}. Some works~\cite{improvingreflow, sd3} hold that a "U-shaped" distribution is optimal and we actually find distribution (0.4, 0.1, 0.1, 0.4) maximized the HPSv2 score. Nevertheless, we found that a front-loaded distribution (0.4, 0.2, 0.2, 0.2) achieved the highest PickScore without a significant compromise in its HPSv2 score. Therefore, we finally adopt the front-loaded strategy and we believe it provides a more balanced and practically advantageous performance.

\subsubsection{Ablations of Key Components}
Finally, to quantify the impact of our core contributions, we start from a baseline model and incrementally add our proposed components: adversarial distillation, Online Trajectory Alignment (OTA), and the improved scheduler. The results in Table~\ref{tab:component_ablation} demonstrate the individual and collective benefits of our method.

The improved scheduler provides the most significant individual performance uplift, highlighting its critical role in few-step generation. Adversarial distillation also delivers a strong boost to the HPSv2 score, confirming its power in enhancing perceptual realism.
While Online Trajectory Alignment (OTA) shows a more modest numerical gain in isolation, we hold it represents the core idea of FlowSteer-\textit{Tarjectory Alignment}.
It directly resolves the training-inference discrepancy in prior work~\cite{perflow, pcm, lcm,improvingreflow}, ensuring the student learns from authentic trajectories rather than artificial interpolated ones. We posit that this principled alignment creates a stable and correct foundation, the powerful synergistic effect observed in our full model validates this hypothesis. It demonstrates that OTA provides the proper on-trajectory data that allows both the adversarial loss and the improved scheduler to function optimally, leading to the overall success of our method.

\section{Discussion}

The core idea of FlowSteer is simple: to make student models learn from the teacher's real generation process, not an artificial one. We first identified previous mrthod's critical flaw leading to both suboptimal guidance from the teacher and error accumulation for the student during inference, and then propose OTA to address it. OTA ensures that the student's training perfectly mirrors its inference, closing the training-inference gap.  
Furthermore, by applying adversarial distillation directly on generation trajectories, we give the student more fine-grained feedback, encouraging the student not just to reach a good destination, but to follow a realistic path to get there.
Finally, our fixed ~\texttt{FlowMatchEulerDiscreteScheduler} yields a surprisingly large performance boost. This underscores the necessity of carefully scrutinizing every component when optimizing for extreme efficiency.\

In conclusion, FlowSteer demonstrates that by systematically aligning the training process with the realities of inference, we can unlock the full potential of ReFlow-based distillation. Through the synergistic combination of on-trajectory data generation (OTA), process-level adversarial feedback, and a corrected sampling scheduler, we have elevated this distillation paradigm to be highly competitive with other state-of-the-art acceleration techniques for modern, large-scale generative models.

\clearpage
{
    \small
    \bibliographystyle{ieeenat_fullname}
    \bibliography{main}

@String(ICLR = {Int. Conf. Learn. Represent.})

@String(AAAI = {AAAI})

@String(ICLR  = {ICLR})

@article{ediff,
  title={ediff-i: Text-to-image diffusion models with an ensemble of expert denoisers},
  author={Balaji, Yogesh and Nah, Seungjun and Huang, Xun and Vahdat, Arash and Song, Jiaming and Zhang, Qinsheng and Kreis, Karsten and Aittala, Miika and Aila, Timo and Laine, Samuli and others},
  journal={arXiv preprint arXiv:2211.01324},
  year={2022}
}

@inproceedings{ufogen,
  title={Ufogen: You forward once large scale text-to-image generation via diffusion gans},
  author={Xu, Yanwu and Zhao, Yang and Xiao, Zhisheng and Hou, Tingbo},
  booktitle={Proceedings of the IEEE/CVF Conference on Computer Vision and Pattern Recognition},
  pages={8196--8206},
  year={2024}
}

@inproceedings{sd3,
  title={Scalable diffusion models with transformers},
  author={Peebles, William and Xie, Saining},
  booktitle={Proceedings of the IEEE/CVF international conference on computer vision},
  pages={4195--4205},
  year={2023}
}

@article{ddpm,
  title={Denoising diffusion probabilistic models},
  author={Ho, Jonathan and Jain, Ajay and Abbeel, Pieter},
  journal={Advances in neural information processing systems},
  volume={33},
  pages={6840--6851},
  year={2020}
}

@article{tian2025audiox,
  title={Audiox: Diffusion transformer for anything-to-audio generation},
  author={Tian, Zeyue and Jin, Yizhu and Liu, Zhaoyang and Yuan, Ruibin and Tan, Xu and Chen, Qifeng and Xue, Wei and Guo, Yike},
  journal={arXiv preprint arXiv:2503.10522},
  year={2025}
}

@article{edm,
  title={Elucidating the design space of diffusion-based generative models},
  author={Karras, Tero and Aittala, Miika and Aila, Timo and Laine, Samuli},
  journal={Advances in neural information processing systems},
  volume={35},
  pages={26565--26577},
  year={2022}
}

@inproceedings{songscore,
  title={Score-Based Generative Modeling through Stochastic Differential Equations},
  author={Song, Yang and Sohl-Dickstein, Jascha and Kingma, Diederik P and Kumar, Abhishek and Ermon, Stefano and Poole, Ben},
  booktitle={International Conference on Learning Representations},
  year={2021}
}

@inproceedings{cm,
  title={Consistency Models},
  author={Song, Yang and Dhariwal, Prafulla and Chen, Mark and Sutskever, Ilya},
  booktitle={International Conference on Machine Learning},
  pages={32211--32252},
  year={2023},
  organization={PMLR}
}

@article{qwenimage,
  title={Qwen-image technical report},
  author={Wu, Chenfei and Li, Jiahao and Zhou, Jingren and Lin, Junyang and Gao, Kaiyuan and Yan, Kun and Yin, Sheng-ming and Bai, Shuai and Xu, Xiao and Chen, Yilei and others},
  journal={arXiv preprint arXiv:2508.02324},
  year={2025}
}

@article{stepvideo,
  title={Step-video-t2v technical report: The practice, challenges, and future of video foundation model},
  author={Ma, Guoqing and Huang, Haoyang and Yan, Kun and Chen, Liangyu and Duan, Nan and Yin, Shengming and Wan, Changyi and Ming, Ranchen and Song, Xiaoniu and Chen, Xing and others},
  journal={arXiv preprint arXiv:2502.10248},
  year={2025}
}

@inproceedings{ladd,
  title={Fast high-resolution image synthesis with latent adversarial diffusion distillation},
  author={Sauer, Axel and Boesel, Frederic and Dockhorn, Tim and Blattmann, Andreas and Esser, Patrick and Rombach, Robin},
  booktitle={SIGGRAPH Asia 2024 Conference Papers},
  pages={1--11},
  year={2024}
}

@inproceedings{2015deep,
  title={Deep unsupervised learning using nonequilibrium thermodynamics},
  author={Sohl-Dickstein, Jascha and Weiss, Eric and Maheswaranathan, Niru and Ganguli, Surya},
  booktitle={International conference on machine learning},
  pages={2256--2265},
  year={2015},
  organization={pmlr}
}

@article{lcm,
  title={Latent consistency models: Synthesizing high-resolution images with few-step inference},
  author={Luo, Simian and Tan, Yiqin and Huang, Longbo and Li, Jian and Zhao, Hang},
  journal={arXiv preprint arXiv:2310.04378},
  year={2023}
}

@inproceedings{yinvideo,
  title={From slow bidirectional to fast autoregressive video diffusion models},
  author={Yin, Tianwei and Zhang, Qiang and Zhang, Richard and Freeman, William T and Durand, Fredo and Shechtman, Eli and Huang, Xun},
  booktitle={Proceedings of the Computer Vision and Pattern Recognition Conference},
  pages={22963--22974},
  year={2025}
}

@inproceedings{lldm,
  title={Large Language Diffusion Models},
  author={Nie, Shen and Zhu, Fengqi and You, Zebin and Zhang, Xiaolu and Ou, Jingyang and Hu, Jun and ZHOU, JUN and Lin, Yankai and Wen, Ji-Rong and Li, Chongxuan},
  booktitle={ICLR 2025 Workshop on Deep Generative Model in Machine Learning: Theory, Principle and Efficacy},
  year={2025}
}

@article{pcm,
  title={Phased consistency models},
  author={Wang, Fu-Yun and Huang, Zhaoyang and Bergman, Alexander and Shen, Dazhong and Gao, Peng and Lingelbach, Michael and Sun, Keqiang and Bian, Weikang and Song, Guanglu and Liu, Yu and others},
  journal={Advances in neural information processing systems},
  volume={37},
  pages={83951--84009},
  year={2024}
}

@inproceedings{makeaudio,
  title={Make-an-audio: text-to-audio generation with prompt-enhanced diffusion models},
  author={Huang, Rongjie and Huang, Jiawei and Yang, Dongchao and Ren, Yi and liu, Luping and Li, Mingze and Ye, Zhenhui and Liu, Jinglin and Yin, Xiang and Zhao, Zhou},
  booktitle={Proceedings of the 40th International Conference on Machine Learning},
  pages={13916--13932},
  year={2023}
}

@inproceedings{pd,
  title={Progressive Distillation for Fast Sampling of Diffusion Models},
  author={Salimans, Tim and Ho, Jonathan},
  booktitle={International Conference on Learning Representations},
  year={2022}
}

@article{dream7b,
  title={Dream 7b: Diffusion large language models},
  author={Ye, Jiacheng and Xie, Zhihui and Zheng, Lin and Gao, Jiahui and Wu, Zirui and Jiang, Xin and Li, Zhenguo and Kong, Lingpeng},
  journal={arXiv preprint arXiv:2508.15487},
  year={2025}
}

@inproceedings{shoutcut,
  title={One Step Diffusion via Shortcut Models},
  author={Frans, Kevin and Hafner, Danijar and Levine, Sergey and Abbeel, Pieter},
  booktitle={The Thirteenth International Conference on Learning Representations},
  year={2025}
}

@inproceedings{icm,
  title={Improved Techniques for Training Consistency Models},
  author={Song, Yang and Dhariwal, Prafulla},
  booktitle={International Conference on Machine Learning},
  year={2024},
}

@inproceedings{ctm,
  title={Consistency Trajectory Models: Learning Probability Flow ODE Trajectory of Diffusion},
  author={Kim, Dongjun and Lai, Chieh-Hsin and Liao, Wei-Hsiang and Murata, Naoki and Takida, Yuhta and Uesaka, Toshimitsu and He, Yutong and Mitsufuji, Yuki and Ermon, Stefano},
  booktitle={ICLR},
  year={2024}
}

@inproceedings{ddim,
  title={Denoising Diffusion Implicit Models},
  author={Song, Jiaming and Meng, Chenlin and Ermon, Stefano},
  booktitle={International Conference on Learning Representations},
  year={2020}
}

@article{fluxreason6m,
  title={Flux-reason-6m \& prism-bench: A million-scale text-to-image reasoning dataset and comprehensive benchmark},
  author={Fang, Rongyao and Yu, Aldrich and Duan, Chengqi and Huang, Linjiang and Bai, Shuai and Cai, Yuxuan and Wang, Kun and Liu, Si and Liu, Xihui and Li, Hongsheng},
  journal={arXiv preprint arXiv:2509.09680},
  year={2025}
}

@inproceedings{dmd,
  title={One-step diffusion with distribution matching distillation},
  author={Yin, Tianwei and Gharbi, Micha{\"e}l and Zhang, Richard and Shechtman, Eli and Durand, Fredo and Freeman, William T and Park, Taesung},
  booktitle={Proceedings of the IEEE/CVF conference on computer vision and pattern recognition},
  pages={6613--6623},
  year={2024}
}

@inproceedings{simplifyingcm,
  title={Simplifying, Stabilizing and Scaling Continuous-time Consistency Models},
  author={Lu, Cheng and Song, Yang},
  booktitle={International Conference on Machine Learning},
  year={2025}
}

@inproceedings{lora,
  title={LoRA: Low-Rank Adaptation of Large Language Models},
  author={Hu, Edward J and Wallis, Phillip and Allen-Zhu, Zeyuan and Li, Yuanzhi and Wang, Shean and Wang, Lu and Chen, Weizhu and others},
  booktitle={International Conference on Learning Representations}
}

@inproceedings{add,
  title={Adversarial diffusion distillation},
  author={Sauer, Axel and Lorenz, Dominik and Blattmann, Andreas and Rombach, Robin},
  booktitle={European Conference on Computer Vision},
  pages={87--103},
  year={2024},
  organization={Springer}
}

@article{improvingreflow,
  title={Improving the training of rectified flows},
  author={Lee, Sangyun and Lin, Zinan and Fanti, Giulia},
  journal={Advances in neural information processing systems},
  volume={37},
  pages={63082--63109},
  year={2024}
}

@inproceedings{flowmatch,
  title={Flow Matching for Generative Modeling},
  author={Lipman, Yaron and Chen, Ricky TQ and Ben-Hamu, Heli and Nickel, Maximilian and Le, Matthew},
  booktitle={International Conference on Learning Representations},
  year={2023}
}

@inproceedings{dcm,
  title={Dual-Expert Consistency Model for Efficient and High-Quality Video Generation},
  author={Lv, Zhengyao and Si, Chenyang and Pan, Tianlin and Chen, Zhaoxi and Wong, Kwan-Yee K and Qiao, Yu and Liu, Ziwei},
  booktitle={Proceedings of the IEEE/CVF International Conference on Computer Vision},
  pages={14983--14993},
  year={2025}
}

@inproceedings{flowstraight,
  title={Flow Straight and Fast: Learning to Generate and Transfer Data with Rectified Flow},
  author={Liu, Xingchao and Gong, Chengyue and Liu, Qiang},
  booktitle={International Conference on Learning Representations (ICLR)},
  year={2023}
}

@article{hypersd,
  title={Hyper-sd: Trajectory segmented consistency model for efficient image synthesis},
  author={Ren, Yuxi and Xia, Xin and Lu, Yanzuo and Zhang, Jiacheng and Wu, Jie and Xie, Pan and Wang, Xing and Xiao, Xuefeng},
  journal={Advances in Neural Information Processing Systems},
  volume={37},
  pages={117340--117362},
  year={2024}
}

@inproceedings{instaflow,
  title={Instaflow: One step is enough for high-quality diffusion-based text-to-image generation},
  author={Liu, Xingchao and Zhang, Xiwen and Ma, Jianzhu and Peng, Jian and others},
  booktitle={International Conference on Learning Representations},
  year={2024}
}

@article{perflow,
  title={Perflow: Piecewise rectified flow as universal plug-and-play accelerator},
  author={Yan, Hanshu and Liu, Xingchao and Pan, Jiachun and Liew, Jun Hao and Liu, Qiang and Feng, Jiashi},
  journal={Advances in Neural Information Processing Systems},
  volume={37},
  pages={78630--78652},
  year={2024}
}

@inproceedings{flash,
  title={Flash diffusion: Accelerating any conditional diffusion model for few steps image generation},
  author={Chadebec, Clement and Tasar, Onur and Benaroche, Eyal and Aubin, Benjamin},
  booktitle={Proceedings of the AAAI Conference on Artificial Intelligence},
  volume={39},
  number={15},
  pages={15686--15695},
  year={2025}
}

@inproceedings{minimizing,
  title={Minimizing trajectory curvature of ode-based generative models},
  author={Lee, Sangyun and Kim, Beomsu and Ye, Jong Chul},
  booktitle={International Conference on Machine Learning},
  pages={18957--18973},
  year={2023},
  organization={PMLR}
}

@article{improving,
  title={Improving the training of rectified flows},
  author={Lee, Sangyun and Lin, Zinan and Fanti, Giulia},
  journal={Advances in neural information processing systems},
  volume={37},
  pages={63082--63109},
  year={2024}
}

@inproceedings{rectifieddiffusion,
  title={Rectified Diffusion: Straightness Is Not Your Need in Rectified Flow},
  author={Wang, Fu-Yun and Yang, Ling and Huang, Zhaoyang and Wang, Mengdi and Li, Hongsheng},
  booktitle={International Conference on Learning Representations},
  year={2025}
}

@article{dmd2,
  title={Improved distribution matching distillation for fast image synthesis},
  author={Yin, Tianwei and Gharbi, Micha{\"e}l and Park, Taesung and Zhang, Richard and Shechtman, Eli and Durand, Fredo and Freeman, Bill},
  journal={Advances in neural information processing systems},
  volume={37},
  pages={47455--47487},
  year={2024}
}

@article{imagineflash,
  title={Imagine flash: Accelerating emu diffusion models with backward distillation},
  author={Kohler, Jonas and Pumarola, Albert and Sch{\"o}nfeld, Edgar and Sanakoyeu, Artsiom and Sumbaly, Roshan and Vajda, Peter and Thabet, Ali},
  journal={arXiv preprint arXiv:2405.05224},
  year={2024}
}

@inproceedings{cfgdistillation,
  title={On distillation of guided diffusion models},
  author={Meng, Chenlin and Rombach, Robin and Gao, Ruiqi and Kingma, Diederik and Ermon, Stefano and Ho, Jonathan and Salimans, Tim},
  booktitle={Proceedings of the IEEE/CVF conference on computer vision and pattern recognition},
  pages={14297--14306},
  year={2023}
}

@article{sdxllightning,
  title={Sdxl-lightning: Progressive adversarial diffusion distillation},
  author={Lin, Shanchuan and Wang, Anran and Yang, Xiao},
  journal={arXiv preprint arXiv:2402.13929},
  year={2024}
}

@article{pickscore,
  title={Pick-a-pic: An open dataset of user preferences for text-to-image generation},
  author={Kirstain, Yuval and Polyak, Adam and Singer, Uriel and Matiana, Shahbuland and Penna, Joe and Levy, Omer},
  journal={Advances in neural information processing systems},
  volume={36},
  pages={36652--36663},
  year={2023}
}

@inproceedings{apt,
  title={Diffusion Adversarial Post-Training for One-Step Video Generation},
  author={Lin, Shanchuan and Xia, Xin and Ren, Yuxi and Yang, Ceyuan and Xiao, Xuefeng and Jiang, Lu},
  booktitle={Forty-second International Conference on Machine Learning},
  year={2025}
}

@article{hpsv2,
  title={Human preference score v2: A solid benchmark for evaluating human preferences of text-to-image synthesis},
  author={Wu, Xiaoshi and Hao, Yiming and Sun, Keqiang and Chen, Yixiong and Zhu, Feng and Zhao, Rui and Li, Hongsheng},
  journal={arXiv preprint arXiv:2306.09341},
  year={2023}
}
}

\clearpage
\setcounter{page}{1}
\maketitlesupplementary

\section{Proof of Inter-Stage Distribution Mismatch}
\label{sec:appendix_mismatch_proof}

Here, we provide the detailed proof that a distribution mismatch occurs at stage boundaries unless the teacher model is a perfect Rectified Flow model.

Let the joint distribution of data and noise be $p(z_0, \epsilon) = p_{\text{data}}(z_0)\mathcal{N}(\epsilon|0,I)$. The condition for a seamless transition between stages requires that the distribution of the teacher's evolved state at time $t_{k-1}$, denoted $\hat{z}_{t_{k-1}}$, matches the prescribed distribution of the ideal starting state $z_{t_{k-1}}$. A necessary condition for this is that their expectations match:
\begin{equation}
    \mathbb{E}_{z_0, \epsilon}[\hat{z}_{t_{k-1}}] = \mathbb{E}_{z_0, \epsilon}[z_{t_{k-1}}].
\end{equation}
The teacher's output $\hat{z}_{t_{k-1}}$ is obtained by integrating its velocity field $v_T$ from a starting point $z_{t_k}$. Taking the expectation over the integral form gives:
\begin{equation}
    \mathbb{E}[z_{t_k}] + \mathbb{E}\left[\int_{t_k}^{t_{k-1}} v_T(z(s), s) ds\right] = \mathbb{E}[z_{t_{k-1}}].
\end{equation}
Using the linearity of expectation and Fubini's theorem to swap the expectation and integral, we get:
\begin{equation}
    \int_{t_k}^{t_{k-1}} \mathbb{E}[v_T(z(s), s)] ds = \mathbb{E}[z_{t_{k-1}}] - \mathbb{E}[z_{t_k}].
    \label{eq:integral_expectation_appendix}
\end{equation}
The expectation of the state $z_t = (1-\sigma(t))z_0 + \sigma(t)\epsilon$ is $\mathbb{E}[z_t] = (1-\sigma(t))\mathbb{E}[z_0] = (1-\sigma(t))\mu_{\text{data}}$, where $\mu_{\text{data}}$ is the mean of the data distribution and $\mathbb{E}[\epsilon]=0$. Substituting this into the right-hand side of Eq.~\ref{eq:integral_expectation_appendix} yields:
\begin{align}
    \mathbb{E}[z_{t_{k-1}}] - \mathbb{E}[z_{t_k}] &= (1-\sigma(t_{k-1}))\mu_{\text{data}} - (1-\sigma(t_k))\mu_{\text{data}} \nonumber \\
    &= (\sigma(t_k) - \sigma(t_{k-1}))\mu_{\text{data}}.
\end{align}
By dividing both sides of Eq.~\ref{eq:integral_expectation_appendix} by $(t_{k-1} - t_k)$ and taking the limit as $t_{k-1} \to t_k$, the left side becomes the instantaneous expected velocity, and the right side becomes the derivative of $-\sigma(t)$ multiplied by $\mu_{\text{data}}$:
\begin{equation}
    \mathbb{E}_{z_t \sim p(z,t)}[v_T(z_t, t)] = -\sigma'(t)\mu_{\text{data}}.
    \label{eq:expected_velocity_appendix}
\end{equation}
This equation represents a powerful constraint. It dictates that for a seamless transition, the mean of the teacher's vector field, when averaged over the entire distribution of states $p(z,t)$ at time $t$, must follow a predetermined path solely defined by the data mean $\mu_{\text{data}}$ and the noise schedule $\sigma(t)$.

However, a general teacher model $v_T$ (e.g., a pre-trained diffusion model) possesses complex, non-linear dynamics and is not guaranteed to satisfy this condition. This condition is only naturally satisfied if the teacher itself is a perfect Rectified Flow model, where the velocity field is specifically constructed to ensure linear trajectories on average. Forcing a general teacher to satisfy this condition contradicts the premise that it has a complex, non-linear trajectory that requires distillation. Therefore, a distribution mismatch is inevitable.

\section{Improved Scheduler Details}
\label{sec:appendix_scheduler}

\definecolor{methodAcolor}{HTML}{D9534F} 
\definecolor{methodBcolor}{HTML}{428BCA} 

\newcommand{\algmark}[2]{%
    \textcolor{#1}{\vrule width 1.2pt height 1.8ex depth 0.2ex \hspace{0.8em}}%
    \texttt{#2}%
}

\begin{algorithm}[t!]
    \caption{sigma sampling logic. Our method (\textcolor{methodBcolor}{blue marker}) samples from a pre-completed schedule, while the original (\textcolor{methodAcolor}{red marker}) appends $\sigma=0$ post-sampling.}
    \label{alg:scheduler_logic}
    \begin{algorithmic}[1]
        \Statex \textit{// --- Original Method ---}
        \State \algmark{methodAcolor}{indices = linspace(0, 999, N)}
        \State \algmark{methodAcolor}{s\_sampled = sigmas[indices]}
        \State \algmark{methodAcolor}{return append(s\_sampled, 0)}
        
        \Statex 
        
        \Statex \textit{// --- Our Improved Method ---}
        \State \algmark{methodBcolor}{s\_full = append(sigmas, 0)}
        \State \algmark{methodBcolor}{indices = linspace(0, 1000, N+1)}
        \State \algmark{methodBcolor}{return s\_full[indices]}
    \end{algorithmic}
\end{algorithm}

\begin{table}[t]
    \centering
    \small 
    \caption{Generated sigma values for N=4 inference steps. }
    \label{tab:sigma_comparison}
    \begin{tabularx}{\columnwidth}{@{} l >{\raggedright\arraybackslash}X @{}}
        \toprule
        \textbf{Setting} & \textbf{Generated Sigma Values} \\
        \midrule
        \multicolumn{2}{@{}l}{\textbf{Original Method}} \\
        \quad Shift=1 & \texttt{[1.000, 0.667, 0.334, 0.001, 0.000]} \\
        \quad Shift=3 & \texttt{[1.000, 0.858, 0.602, 0.009, 0.000]} \\
        \addlinespace 
        \multicolumn{2}{@{}l}{\textbf{Our Improved Method}} \\
        \quad Shift=1 & \texttt{[1.000, 0.750, 0.500, 0.250, 0.000]} \\
        \quad Shift=3 & \texttt{[1.000, 0.900, 0.751, 0.502, 0.000]} \\
        \bottomrule
    \end{tabularx}
\end{table}

As discussed, \texttt{FlowMatchEulerDiscreteScheduler} exhibits a structural flaw in the few-step inference. The original implementation samples $N$ sigmas from the full schedule and then appends the terminal state ($\sigma=0$) separately. This results in a final step size that is not proportional to the preceding steps, which can degrade image quality.

To rectify this, we propose an improved sampling strategy. Our method first augments the full sigma schedule with the terminal $\sigma=0$ state. Then, it samples $N+1$ points from this complete schedule. This ensures that all step intervals are proportionally scaled.

The unified logic, highlighting the divergence between the original and our improved method, is presented in Algorithm~\ref{alg:scheduler_logic}. A concrete example of the resulting sigma values for a 4-step inference is shown in Table~\ref{tab:sigma_comparison}.

\end{document}